\documentclass[10pt,twocolumn,letterpaper]{article}

\usepackage{cvpr}              

\usepackage{multirow}
\usepackage{color,xcolor}
\usepackage{pifont}
\usepackage{makecell}
\newcommand{\cmark}{\ding{51}} 
\newcommand{\xmark}{\ding{55}} 

\definecolor{table_red}{RGB}{255,46,99}
\definecolor{table_blue}{RGB}{8,217,214}
\definecolor{table_gray}{RGB}{156,156,156}

\definecolor{cvprblue}{rgb}{0.21,0.49,0.74}
\usepackage[pagebackref,breaklinks,colorlinks,allcolors=cvprblue]{hyperref}

\def\paperID{} 
\def\confName{CVPR}
\def\confYear{2026}

\title{UmniBench: Unified Understanding and Generation \\ Model Oriented Omni-dimensional Benchmark}

\author{
  Kai Liu$^{1}$\thanks{Equal contribution}~,\enspace
  Leyang Chen$^{1}$\footnotemark[1]~,\enspace
  Wenbo Li$^{2}$,\enspace
  Zhikai Chen$^{3}$,\enspace
  Zhixin Wang$^{3}$,\enspace \\
  Renjing Pei$^{3}$,\enspace
  ~Linghe Kong$^{1}$\footnotemark[2]~,\enspace
  Yulun Zhang$^{1}$\thanks{Corresponding authors: Yulun Zhang, yulun100@gmail.com, Linghe Kong,  linghe.kong@sjtu.edu.cn}\\
  \textsuperscript{1}Shanghai Jiao Tong University,\enspace
  \textsuperscript{2}The Chinese University of Hong Kong,\enspace 
  \textsuperscript{3}Huawei Technologies Ltd.\enspace \\
  Homepage:~\hyperlink{https://umnibench.github.io/}{https://umnibench.github.io/}
}

\begin{document}
\maketitle

\setlength{\abovedisplayskip}{2pt}
\setlength{\belowdisplayskip}{2pt}

\begin{abstract}
Unifying multimodal understanding and generation has shown impressive capabilities in cutting-edge proprietary systems.
However, evaluations of unified multimodal models (UMMs) remain decoupled, assessing their understanding and generation abilities separately with corresponding datasets. 
To address this, we propose UmniBench, a benchmark tailored for UMMs with omni-dimensional evaluation.
First, UmniBench can assess the understanding, generation, and editing ability within a single evaluation process.
Based on human-examined prompts and QA pairs, UmniBench leverages UMM itself to evaluate its generation and editing ability with its understanding ability.
This simple but effective paradigm allows comprehensive evaluation of UMMs. 
Second, UmniBench covers 13 major domains and more than 200 concepts, ensuring a thorough inspection of UMMs.
Moreover, UmniBench can also decouple and separately evaluate understanding, generation, and editing abilities, providing a fine-grained assessment.
Based on UmniBench, we benchmark 24 popular models, including both UMMs and single-ability large models. 
We hope this benchmark provides a more comprehensive and objective view of unified models and logistical support for improving the performance of the community model.
\end{abstract}
\vspace{-4mm}
\section{Introduction}\label{sec:intro}
A unified multimodal model (UMM) refers to a single model that seamlessly integrates understanding, generation, and editing within a single architecture, with the aim that these capabilities mutually reinforce one another~\cite{deng2025emerging,wu2025omnigen2,wang2025ovis}.
Recent advances in UMMs have shown impressive capabilities in cutting-edge models, which mainly focus on architecture design, post-training paradigms, and dataset construction.
However, current evaluation protocols for these models are largely decoupled~\cite{ghosh2023geneval,wu2024conceptmix}. 
They assess understanding, generation, and editing separately rather than in an integrated manner.
To be specific, understanding benchmarks typically take text and images as input and use multiple-choice answers to probe comprehension and reasoning~\cite{das2024exams}.
By contrast, generation and editing benchmarks take short textual prompts or text–image pairs as input to assess quality and consistency of the output image~\cite{huang2025t2i}.
Despite all these benchmarks allowing accurate evaluation in their corresponding sub-tasks, they are inherently unsuitable for evaluating UMMs.
This is because these evaluations capture each facet in isolation, and they do not reflect the holistic competence of unification, namely, the seamless integration of understanding, generation, and editing.

\begin{figure}[t]
    \centering
    \vspace{-2mm}
    \includegraphics[width=\linewidth]{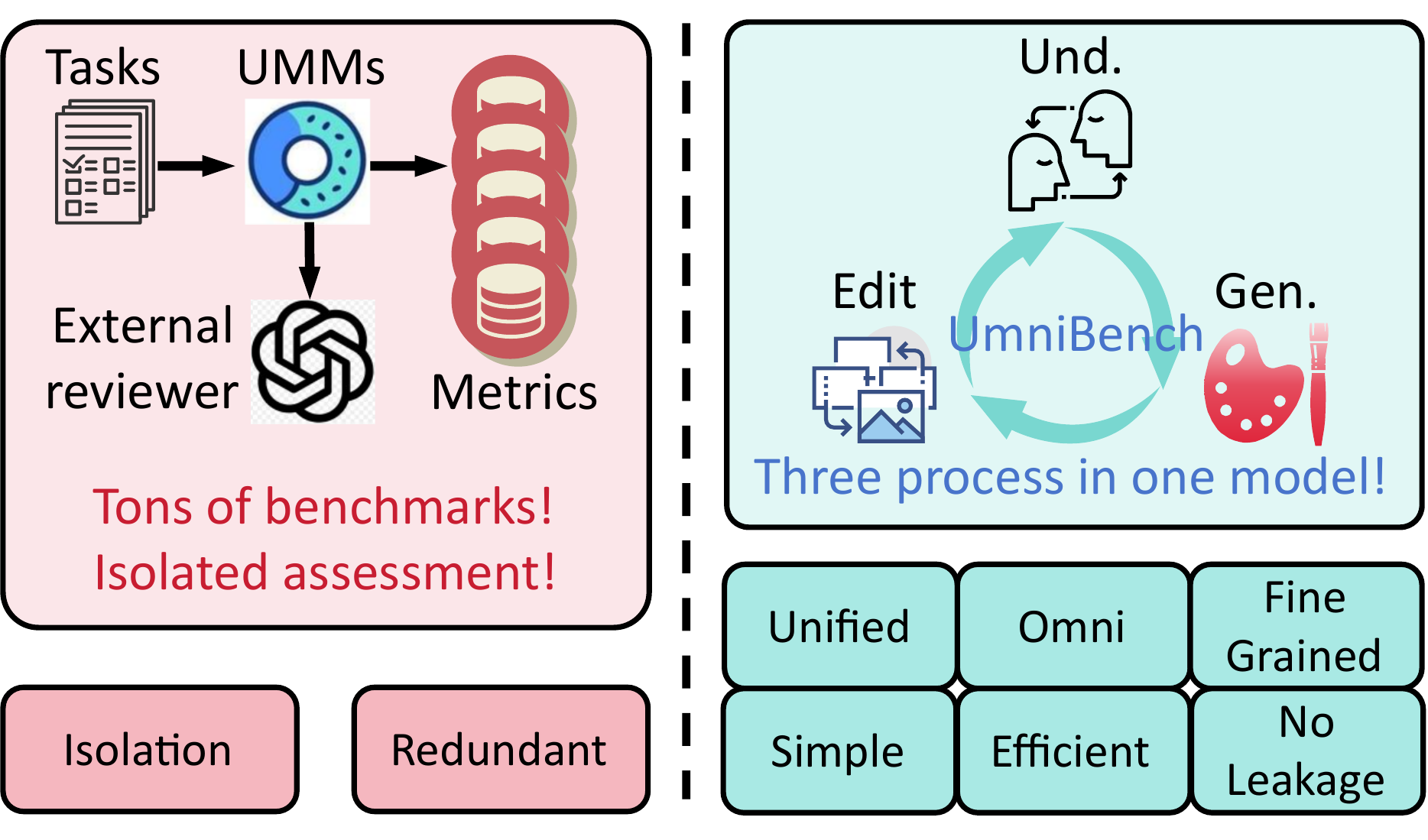}
    \vspace{-8mm}
    \caption{Advantages of our proposed UmniBench compared with previous UMMs isolated evaluation protocols.}
    \vspace{-7mm}
    \label{fig:advantage}
\end{figure}

\begin{figure*}
    \centering
    \includegraphics[width=\linewidth]{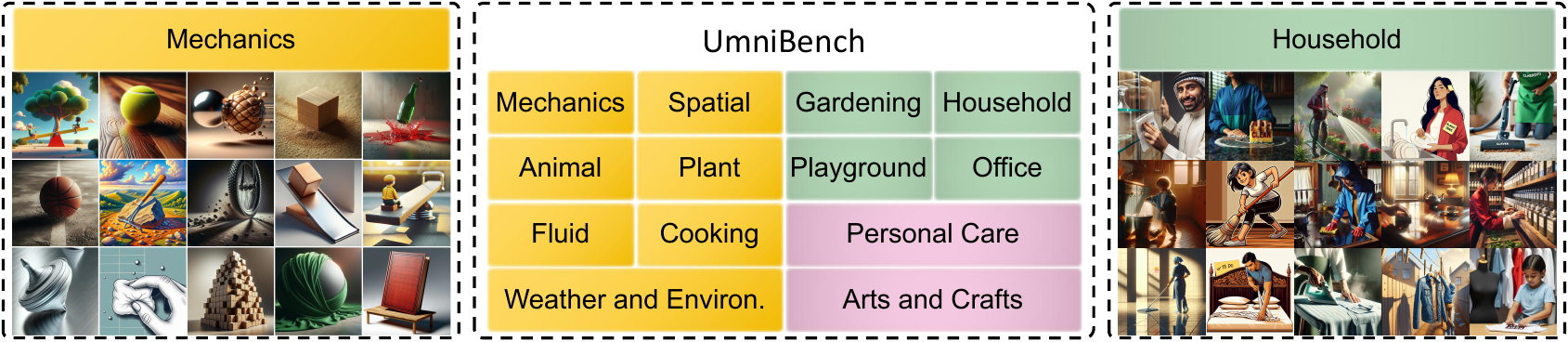}
    \vspace{-6mm}
    \caption{\textbf{The overview of UmniBench.} All 13 domains involved in UmniBench are enumerated in the table above, with left-hand and right-hand panels presenting representative images generated under each concept under the specific domain.}
    \label{fig:overview}
    \vspace{-6mm}
\end{figure*}

Consequently, there is an urgent need for a benchmark specifically designed to evaluate the holistic capabilities of unified generation–understanding models. 
To address this gap, we propose UmniBench, a benchmark tailored for UMMs with omni-dimensional evaluation. 
While existing pure generation and editing models have achieved remarkable image quality, they remain deficient in complex instruction following~\cite{brooks2023instructpix2pix}.
In particular, these models often struggle to accurately interpret complex user intents, especially when the input requires nontrivial reasoning. 
Researchers have therefore begun integrating understanding capabilities into generative models, aiming to improve unified models' handling of complex intents in generation and editing tasks. 
Accordingly, our proposed UmniBench focuses on image generation and editing scenarios that demand substantive understanding and reasoning.

Conventional benchmarks typically rely on external models, hand-crafted rules, or human evaluators to score system outputs~\cite{pan2025ice}. Because UMMs intrinsically possess understanding, generation, and editing capabilities, they can, in principle, serve as their own evaluators.
A natural property of image generation and editing tasks is that the prompt itself constitutes a relatively accurate description of the target image. 
Consequently, we can derive question–answer pairs about image content directly from the prompt and skip the generated image. 
Once a generated or edited image is obtained, these QA pairs can be used to assess: 
(1) whether the model’s generation or editing adheres to the instruction, and (2) whether the model can comprehend the image and the question to produce the correct answer. 
The entire evaluation pipeline can thus be executed by the UMMs alone, enabling self-assessment of its understanding, generation, and editing abilities.

\begin{table*}[t]
\centering
\setlength{\tabcolsep}{10pt}
\caption{\textbf{Summary of Multimodal Benchmarks.} We compare existing works from the following aspects: multi-modal evaluation, number of single abilities, unified ability, dependency on third-party review, and diversity of evaluation.}
\label{tab:benchmark-comp}
\vspace{-3mm}
\resizebox{\linewidth}{!}{
\begin{tabular}{rrcccccc}
\toprule[0.15em]
\multicolumn{1}{c}{\multirow{2}{*}{Benchmark}} & \multicolumn{1}{c}{\multirow{2}{*}{Publication}} & \multirow{2}{*}{Reasoning} & \multicolumn{1}{c}{\multirow{1}{*}{ \# Single}} & \multicolumn{1}{c}{\multirow{1}{*}{Unified}} & \multicolumn{1}{c}{\multirow{1}{*}{No third-party}} & \multicolumn{2}{c}{Diversity} \\
\multicolumn{1}{c}{} & \multicolumn{1}{c}{} &  & \multicolumn{1}{c}{Abilities} & \multicolumn{1}{c}{Ability} & \multicolumn{1}{c}{ review req.} & \multicolumn{1}{c}{Tags} & \multicolumn{1}{c}{Sub-tags} \\
\midrule[0.05em]
\midrule[0.05em]
ReasonPix2Pix~\cite{jin2024reasonpix2pix} & \textcolor{gray}{{\small arXiv'24}}   & \textcolor{table_red}{\cmark} & Edit & \textcolor{table_blue}{\xmark}  & \textcolor{table_red}{\cmark}  & -  & -  \\
ReasonEdit~\cite{huang2024smartedit}      & \textcolor{gray}{{\small arXiv'24}}   & \textcolor{table_red}{\cmark} & Edit & \textcolor{table_blue}{\xmark}  & \textcolor{table_blue}{\xmark} & -  & -  \\
GenEval~\cite{ghosh2023geneval}           & \textcolor{gray}{{\small NeurIPS'23}} & \textcolor{table_blue}{\xmark}& Gen. & \textcolor{table_blue}{\xmark}  & \textcolor{table_blue}{\xmark} & 6  & 6  \\
DPG-Bench~\cite{hu2024ella}               & \textcolor{gray}{{\small arXiv'24}}   & \textcolor{table_blue}{\xmark}& Gen. & \textcolor{table_blue}{\xmark}  & \textcolor{table_blue}{\xmark} & 5  & 13 \\
T2I-CompBench++~\cite{huang2025t2i}       & \textcolor{gray}{{\small NeurIPS'23}} & \textcolor{table_blue}{\xmark}& Gen. & \textcolor{table_blue}{\xmark}  & \textcolor{table_blue}{\xmark} & 4  & 8  \\
ConceptMix~\cite{wu2024conceptmix}        & \textcolor{gray}{{\small NeurIPS'24}} & \textcolor{table_blue}{\xmark}& Gen. & \textcolor{table_blue}{\xmark}  & \textcolor{table_blue}{\xmark} & 8  & 8  \\
KRIS-Bench~\cite{wu2025kris}              & \textcolor{gray}{{\small arXiv'24}}   & \textcolor{table_red}{\cmark} & Edit & \textcolor{table_blue}{\xmark}  & \textcolor{table_blue}{\xmark} & 3  & 22 \\
EditWorld~\cite{zeng2025editworld}        & \textcolor{gray}{{\small MM’25}}      & \textcolor{table_red}{\cmark} & Edit & \textcolor{table_blue}{\xmark}  & \textcolor{table_blue}{\xmark} & 7  & 7  \\
RISEBench~\cite{zhao2025envisioning}      & \textcolor{gray}{{\small NeurIPS'25}} & \textcolor{table_red}{\cmark} & Edit & \textcolor{table_blue}{\xmark}  & \textcolor{table_blue}{\xmark} & 4  & 4  \\
\midrule[0.05em]
UmniBench       & {\textbf{Ours}}               & \textcolor{table_red}{\cmark}             & Und., Gen., Edit    & \textcolor{table_red}{\cmark}   & \textcolor{table_red}{\cmark}  & 13 & 195 \\
\bottomrule[0.15em]
\end{tabular}
} 
\vspace{-5mm}
\end{table*}

Moreover, prior benchmarks typically cover only a single vertical domain of knowledge, such as mathematics or coding. 
In contrast, researchers envision UMMs that span the vast majority of human knowledge. 
Accordingly, UmniBench comprises 13 domains that encompass common categories, including the spatial relations, logical reasoning, and everyday scenarios, as depicted in Fig.~\ref{fig:overview}. 
Within each domain, there are 15 concepts, and each concept contains 3 cases.
Collectively, this design ensures a comprehensive, multi-scenario evaluation of model robustness and stability.
Moreover, the self-generation-self-evaluate paradigm is inherently immune to the risk of data leakage.
Our core contributions can be summarized as threefold:
\begin{itemize}
    \item We propose \textbf{UmniBench}, a benchmark target for UMMs with omni-dimensional evaluation.
    To the best of our knowledge, UmniBench is the first benchmark to assess models not only their understanding and generation abilities individually, but also jointly.
    
    \item UmniBench provides omni-dimensional evaluation.
    13 domains and 15 concepts within each domain are included to conduct a nuanced evaluation.
    Moreover, the unique evaluation method of UmniBench is inherently immune to the risk of data leakage.

    \item State-of-the-art open-sourced methods and proprietary models with unified ability or single ability are evaluated based on our proposed benchmark.
\end{itemize}
\section{Related Works}
\vspace{-1mm}
\subsection{Unified Multimodal Model}
\vspace{-1mm}
UMMs refer to architectures that possess both understanding and generation capabilities, with the aim that the two abilities complement and reinforce each other~\cite{li2025llava,wu2025omnigen2}.
In terms of design philosophy, current designs mainly embed the two capabilities in distinct parameter sets~\cite{li2022blip,li2025llava}.

Accordingly, we can categorize models by their degree of coupling into tightly coupled and loosely coupled variants. 
In \textbf{tightly coupled models}, information flows bidirectionally and interleaves across the two parameter spaces~\cite{deng2025emerging}.
Taking Bagel as an example, a Mixture-of-Experts architecture processes tokens for understanding and generation separately, while allowing interaction through self-attention.

By contrast, \textbf{loosely coupled models} permit primarily unidirectional information flow—from the parameters associated with one ability to those of the other. 
Concretely, in UniPic2, tokens produced by the MLLM for understanding are injected into the diffusion model via a lightweight linear layer.
However, tokens from the diffusion process do not influence the tokens output by the MLLM.
\vspace{-1mm}
\subsection{Benchmarks}
\vspace{-1mm}
Evaluation of UMMs is typically conducted by employing multiple benchmarks that assess each capability in isolation. 
For understanding, commonly used datasets include MME~\cite{fu2025mme}, MMBench~\cite{liu2024mmbench}, MMMU~\cite{yue2024mmmu}, MMVet~\cite{yu2024mmvet}, MathVista~\cite{lu2024mathvista}, and MMVP~\cite{zhang2024mmvp}.
These datasets take text or image–text pairs as inputs and use multiple-choice or numeric answers to evaluate a model’s comprehension and reasoning.
For generations, representative datasets include GenEval~\cite{ghosh2023geneval} and WISE~\cite{niu2025wise}. 
Prompts in these datasets typically specify object attributes, such as spatial position, texture, and quantity, and the evaluation measures the consistency and quality of the generated image.
For editing, benchmarks such as GEdit-Bench~\cite{liu2025step1x} and IntelligentBench~\cite{deng2025emerging} primarily involve real-world images and cover operations such as adding, deleting, replacing, and so on. 
Above evaluation commonly relies on external assessment models heavily, such as GPT-4o~\cite{openai-4o-image-2025}, whose usage is expensive and often limited in many areas.

\section{UmniBench}\label{sec:method}
\vspace{-1mm}
This section introduces UmniBench from two perspectives: the evaluation protocol and the benchmark construction.
\subsection{Evaluation Pipeline}\label{sec:method-evaluation}
\begin{figure*}[t]
    \centering
    \includegraphics[width=\linewidth]{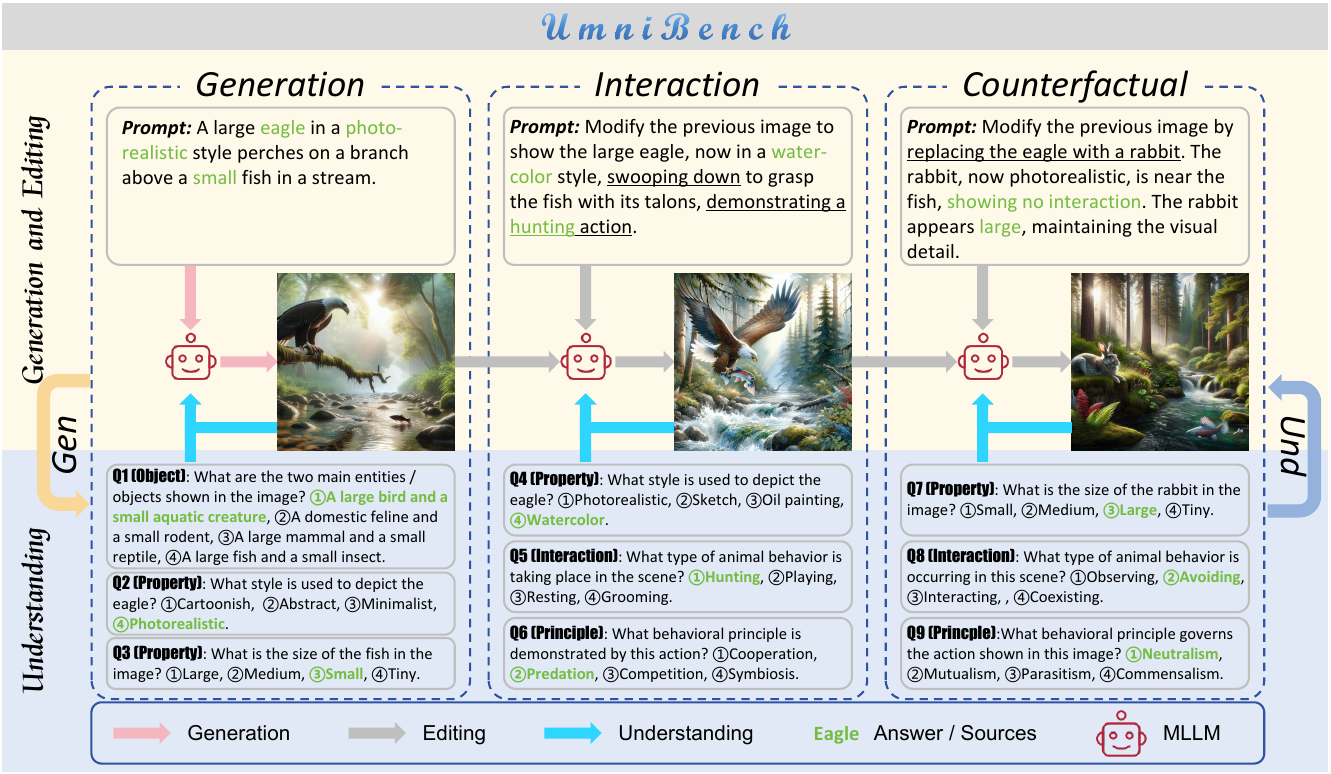}
    \vspace{-6mm}
    \caption{The evaluation pipeline. There are three stages, including generation, interaction, and counterfactual stages. For each stage, there are two parts, \ie, the generation and the understanding part. UMMs take the prompt and image (if it exists) from the previous stage as input and generate a new image. Then, UMMs will be asked questions about the image to evaluate if the image follows the prompt. 
    }
    \label{fig:evaluation}
    \vspace{-5mm}
\end{figure*}

The evaluation pipeline comprises three stages, each consisting of generation and evaluation, as shown in Fig.~\ref{fig:evaluation}.

\noindent\textbf{Generation}. 
The UMM faithfully generates an image according to a specified prompt. 
The prompt includes the spatial relation between a pair of entities as well as detailed attributes of each entity, \eg, color, material, and texture. 
The UMM is then presented with three questions about the image, probing the generation consistency.

\noindent\textbf{Interaction}. 
Building on the first image, the model is required to enact concept-level interactions, \eg, object collisions, predator–prey dynamics, while also applying fine-grained attribute modifications to entities, such as changes in color or texture.
The UMM again receives three questions, assessing if the UMM understands the interaction between the entities and the modifications.
To prevent the model from bypassing visual grounding and answering purely from common sense or the prompt, the attribute edits in this stage typically contradict prior knowledge, for example, changing a lion’s mane to red.

\noindent\textbf{Counterfactual}.
The model is required to perform counterfactual reasoning in this stage.
Concretely, the original entity pair in the image is replaced with a different pair, accompanied by detailed attribute specifications.
The UMM is asked to predict the interaction outcome under the same conceptual setting.
For instance, if the initial scenario depicts a lion hunting a zebra, replacing the pair with an elephant and a zebra yields peaceful coexistence. 
The UMM must again answer three questions based on the updated image, primarily targeting the new attributes and the new interaction results with counterfactual reasoning.

Such a self-generate-self-evaluate process can be inherently immune to the risk of data leakage, avoiding maliciously boosting scores on the leaderboard.

\subsection{Construction Pipeline}
\label{sec:method-construction}
\vspace{-1mm}
The dataset is constructed in a top-down manner, progressively determining the domain, concept, entity set, case, and QA pairs. 
Finally, advanced LLMs and human experts jointly filter out unsuitable test items.

\noindent\textbf{Domains and Concepts.} 
Experts first survey the key competencies of UMMs and propose 13 common domains (see Fig.~\ref{fig:overview}). 
Within each domain, an advanced LLM proposes multiple concepts that cover representative scenarios. 
Experts then filter concepts ill-suited for generation and editing, resulting in a finalized set of 15 concepts.

\noindent\textbf{Entity Pairs and Cases.} 
For each concept, we leverage an LLM to generate multiple entity pairs together with counterfactual pairs.
A counterfactual pair differs from the original in exactly one entity.
The original and counterfactual together constitute a single case. 
For example, if the entity pair is “eagle and fish”, a possible counterfactual pair is “rabbit and fish”, with one entity replaced. 
Next, for each case, the LLM produces prompts corresponding to the three evaluation stages. 
The generation stage prompt focuses on the initial scene and attributes of the entity pair. 
The interaction stage prompt specifies a causal interaction between the entities and a fine-grained attribute change for one entity. 
The counterfactual stage prompt mentions an entity replacement and an additional attribute modification, inspecting the counterfactual reasoning ability of UMMs. 
We also consider a rich attribute space to achieve omni-dimensional evaluation, including color, texture, size, physical properties, position, style, text, facial expression, and more.

\noindent\textbf{Question and Answer Pairs.}
We generate three questions per stage, for nine questions total, and filter out inadequate questions. 
The nine questions are structured across three stages, with specific functional orientations for each query.

The generation stage comprises questions 1 to 3, focusing on entity identification and verification of two core entity attributes.
Q1 serves as the initial scenario inquiry.
Q2 and Q3 are about initial details, inspecting the understanding of details, and consistency between the output image and the prompts.
The interaction stage includes questions 4 to 6, encompassing one attribute-change verification and two causal-interaction inferences.
Q4 is the first post-editing detail-focused question.
Q5 and Q6 address the interaction outcome and principal inquiry, respectively.
The counterfactual stage consists of questions 7 to 9, involving one additional attribute-change verification and two counterfactual inferences.
Q7 functions as the detail-focused question.
Q8 and Q9 explore counterfactual interaction outcomes and principal inquiry, respectively.

Figure~\ref{fig:evaluation} demonstrates an example case with three stages and nine questions.
In aggregate, five questions are perception-oriented—probing semantic and low-level understanding and verification of visual content.
While the remaining four assess reasoning over the entity and the counterfactual entity pair.
One key principle for assessing is that the model can not obtain the answer directly from the question, let alone from common sense.
To avoid such skipping, we adopt abstract terminology that necessitates reliance on images for reasoning questions.
To avoid ambiguity, we explicitly name entities in detail-oriented questions and often set attributes contrary to common sense. 

\noindent\textbf{Validation.}
Validation proceeds with an initial LLM-based coarse filter followed by expert fine-grained review. 
For the coarse filter, we employ the following rules: (1) the answer must not be recoverable from the question alone, (2) options must be semantically unambiguous, and (3) content must exclude violence, gore, or other inappropriate material. 
For the five perception-oriented questions, if three or more violate these rules, the entire case is rejected.
The detailed prompt is shown in the supplementary material.
Subsequently, several experts review the remaining cases with the real outputs from target models and remove unsuitable cases.
An example of an unsuitable case is to demonstrate the electromagnetic force between two stationary objects, which can not be seen with images.

Through the above four-stage construction pipeline, we obtain a high-quality UmniBench that enables comprehensive and reliable evaluation of UMMs.

\begin{figure*}[t]
    \centering
    \includegraphics[width=\linewidth]{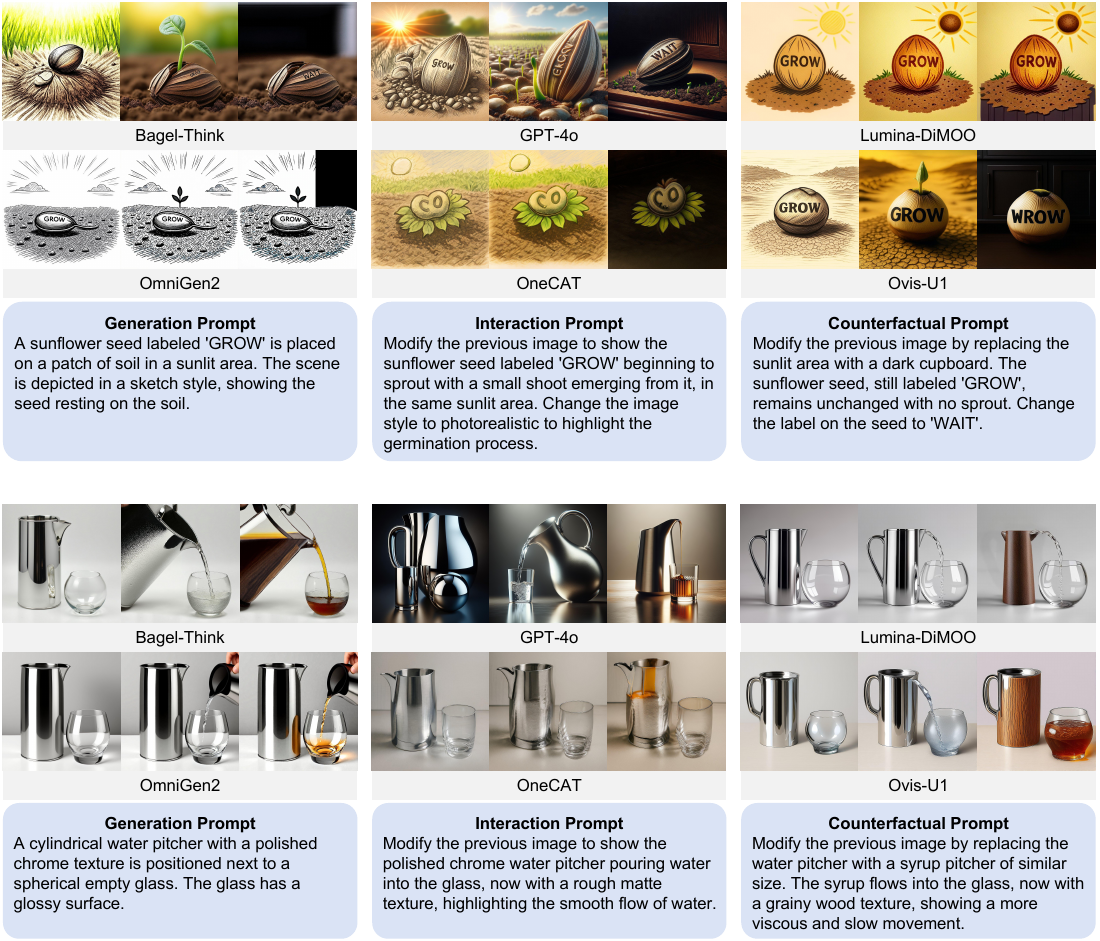}
    \vspace{-3mm}
    \caption{An example of two cases in our proposed UmniBench.
    The UMMs generate or edit the image from the previous stage with specified generation, interaction, and counterfactual prompt.
    The images are placed in sequence while the prompts are provided below.
    Most of the models can provide a successful image in the generation stage, but fail on the counterfactual prompt.}
    \label{fig:display}
    \vspace{-3mm}
\end{figure*}

\subsection{Single Abilities}
UmniBench also supports evaluating individual capabilities. 
Specifically, model invocations within UmniBench fall into three categories: understanding, generation, and editing. 
To assess a model’s performance on one or two capabilities, the other invocation can simply be replaced with corresponding state-of-the-art models.
The reason for the SOTA model determination can be found in the supplementary material.
Therefore, our proposed UmniBench can evaluate large models with a single ability or multiple abilities.

\begin{table*}[t]
\centering
\small
\setlength{\tabcolsep}{2pt}
\caption{Evaluation results of eight advanced UMMs on UmniBench. UmniBench can be divided into three turns, with three questions under each. We calculated the average accuracy of model answers under each question. The definition of Q1 to Q9 is explained in Sec.\ref{sec:method-construction}.}
\label{tab:main-comp}
\vspace{-3mm}
\resizebox{\linewidth}{!}{
\begin{tabular}{l|l|cccccccc}
\toprule[0.15em]
\multicolumn{2}{c}{Model}       & Bagel-Think~\cite{deng2025emerging}{} & Ovis-U1~\cite{wang2025ovis} & Bagel~\cite{deng2025emerging} & UniPic2~\cite{wei2025skywork} & GPT4o~\cite{openai-4o-image-2025} & OmniGen2~\cite{wu2025omnigen2} & OneCAT~\cite{li2025onecat}& Lumina-DiMOO~\cite{xin2025lumina} \\
\midrule[0.05em]
\midrule[0.05em]
\multirow{4}{*}{Generation}                     & Q1      & 97.56       & 98.50   & 97.37 & 95.50   & 97.19 & 96.81    & 94.75  & 88.37        \\
                                           & Q2      & 84.18       & 89.84   & 83.98 & 81.45   & 78.52 & 80.27    & 83.59  & 73.83        \\
                                           & Q3      & 77.11       & 81.74   & 74.39 & 76.02   & 69.75 & 73.84    & 66.21  & 61.85        \\
                                           & Ave. & 87.39       & 91.01   & 86.54 & 85.34   & 83.29 & 84.84    & 83.29  & 76.20        \\
\midrule[0.05em]
\multirow{4}{*}{Interaction}                     & Q4      & 83.11       & 80.86   & 71.67 & 81.61   & 74.30 & 39.96    & 39.96  & 39.96        \\
                                           & Q5      & 81.61       & 77.86   & 81.05 & 80.49   & 85.37 & 72.05    & 65.67  & 46.34        \\
                                           & Q6      & 74.50       & 72.51   & 76.72 & 75.39   & 82.48 & 68.29    & 59.20  & 40.80        \\
                                           & Ave. & 80.03       & 77.32   & 76.47 & 79.37   & 80.62& 59.66    & 54.71  & 42.45        \\
\midrule[0.05em]
\multirow{4}{*}{Counterfactual}                   & Q7      & 70.73       & 69.04   & 62.66 & 58.91   & 63.41 & 37.15    & 41.46  & 35.65        \\
                                           & Q8      & 67.92       & 66.79   & 66.04 & 63.98   & 69.61 & 56.29    & 50.28  & 45.22        \\
                                           & Q9      & 59.62       & 56.53   & 59.38 & 57.48   & 64.13 & 48.93    & 44.66  & 33.49        \\
                                           & Ave. & 66.58       & 64.69   & 62.95 & 60.32   & 65.83& 47.34    & 45.53  & 38.47        \\
\midrule[0.05em]
                                           & Ave. & 77.85       & 77.45   & 75.14 & 74.86   & 76.49 & 63.56    & 60.76  & 51.90        \\
\bottomrule[0.15em]
\end{tabular}
} 
\vspace{-5mm}
\end{table*}
\section{Experiments}
\vspace{-2mm}
\subsection{Implementation}
\vspace{-2mm}
\noindent\textbf{Evaluated Models.}
For unified models, we evaluated the following models: Bagel~\cite{deng2025emerging}, Ovis-U1~\cite{wang2025ovis}, GPT-4o~\cite{openai-4o-image-2025}, UniPic2~\cite{wei2025skywork}, OmniGen2~\cite{wu2025omnigen2}, OneCAT~\cite{li2025onecat}, and Lumina-DiMOO~\cite{xin2025lumina}.
For understanding models, we include Kimi-VL-A3B-Instruct~\cite{team2025kimi}, Janus-pro~\cite{chen2025janus}, QWen3-VL-8B-Instruct~\cite{brooks2023instructpix2pix}, Emu3~\cite{wang2024emu3}, Janus~\cite{wu2025janus}, and BLIP-VQA~\cite{li2022blip}.
For editing models, we include Google Gemini 2.5 Flash Image, QWen-Image-Edit~\cite{wu2025qwen}, Flux.1-Fill~\cite{flux2024}, InstructPix2Pix~\cite{brooks2023instructpix2pix}, MagicBrush~\cite{zhang2023magicbrush}, and Step1X-Edit~\cite{liu2025step1x}.
For generation models, we tested UniPic2~\cite{wei2025skywork}, OneCAT~\cite{li2025onecat}, Emu3~\cite{wang2024emu3}, Janus-Pro~\cite{chen2025janus}, PixArt-$\alpha$~\cite{chen2023pixart}, and Janus~\cite{wu2025janus}.

\noindent\textbf{UmniBench.}
The evaluation pipeline and dataset construction workflow have been elaborated in Sec.~\ref{sec:method}.
For benchmark development and validation, an advanced LLM is employed as the LLM to generate concepts, cases, prompts, validation, and coarse filtering.
Additional details regarding the prompt engineering specifications are provided in the supplementary material.
We also organized five human experts to thoroughly examine the remaining items.
For the single-ability assessment, we selected task-specific state-of-the-art (SOTA) models to isolate individual capabilities.
QWen3-VL is adopted for the understanding task
and Flux.1 Fill for image editing.
The reason for the selection can be found in the supplementary material.

\subsection{Unified Ability}
The results of UMMs on UmniBench along generation, interaction, and counterfactual stages are shown in Tab.~\ref{tab:main-comp}.

\noindent\textbf{Hierarchical Stratification}.
Bagel-Think achieves the highest aggregate performance (77.85) while maintaining the most stable degradation (87.39 → 80.03 → 66.58). 
Ovis-U1 attains comparable overall results (77.45) but manifests greater volatility across stages.
Whereas Lumina-DiMOO's score (51.90) represents only 66.6\% of the leading performance, indicating a substantial capability gap.

\noindent\textbf{Monotonic Decrement on Stages.}
All models exhibit a consistent monotonic decrement across evaluation stages, with mean scores declining from 83.99 (Generation) to 72.76 (Interaction) and 59.55 (Counterfactual)—a cumulative reduction of 29.1\%.
This uniform decline pattern substantiates the effectiveness of the designed difficulty gradient and reveals systematic challenges in long-horizon reasoning and adaptive capacity across all evaluated systems.

\noindent\textbf{Stability Analysis.}
Stability analysis reveals marked divergences.
Bagel-Think and Ovis-U1 maintain relatively low inter-turn variance ($SD \approx 10.4$), while models such as OmniGen2, OneCAT, and Lumina-DiMOO exhibit catastrophic performance degradation in later stages. 
Notably, OmniGen2 experiences a 44.2\% relative decline from the generation stage to the counterfactual stage, evidencing severe robustness deficits when confronted with counterfactual reasoning scenarios with high complexity.

\begin{table*}[t]
\centering
\setlength{\tabcolsep}{4pt}
\caption{Evaluation results under each domain. We calculate the average accuracy of model responses across each domain. Horizontal comparison of these domain scores enables quantitative assessment of models' distinct performance within a particular domain.}
\label{tab:comp-uni-domain}
\vspace{-3mm}
\resizebox{\linewidth}{!}{
\begin{tabular}{c|cccccccc|c}
\toprule[0.15em]
Domain                  & Bagel-think~\cite{deng2025emerging} & Ovis-U1~\cite{wang2025ovis} & Bagel~\cite{deng2025emerging} & UniPic2~\cite{wei2025skywork} & GPT4o~\cite{openai-4o-image-2025} & OmniGen2~\cite{wu2025omnigen2} & OneCAT~\cite{li2025onecat}& Lumina-DiMOO~\cite{xin2025lumina} & Average \\
\midrule[0.05em]
\midrule[0.05em]
Spatial                 & 63.16       & 62.54   & 62.85 & 57.28   & 56.66 & 49.54    & 55.11  & 47.68        & 57.04   \\
Plant                   & 70.99       & 67.94   & 59.92 & 66.41   & 69.08 & 56.87    & 49.24  & 42.75        & 59.88   \\
Fluid                   & 68.05       & 70.41   & 67.46 & 67.16   & 70.12 & 50.00    & 53.55  & 46.45        & 61.76   \\
Physical                & 70.16       & 68.89   & 69.21 & 67.62   & 96.84 & 62.54    & 56.51  & 51.43        & 63.17   \\
Cooking                 & 75.32       & 74.68   & 76.28 & 72.12   & 73.40 & 64.74    & 60.90  & 43.59        & 67.83   \\
Arts and Crafts         & 78.32       & 77.17   & 73.70 & 70.81   & 76.59 & 62.72    & 54.05  & 48.55        & 67.88   \\
Weather \& Environ.    & 72.73       & 77.27   & 72.16 & 71.88   & 75.28& 58.52    & 68.47  & 55.68        & 68.75   \\
Animal                  & 77.43       & 79.79   & 76.12 & 79.79   & 79.27 & 57.74    & 63.25  & 47.77        & 69.13   \\
Office                  & 82.34       & 78.44   & 77.54 & 78.74   & 81.73 & 68.56    & 61.08  & 56.29        & 72.57   \\
Personal Care           & 84.75       & 81.69   & 80.00 & 79.66   & 86.10 & 73.56    & 69.83  & 53.90        & 75.93   \\
Playground              & 82.32       & 84.60   & 81.82 & 83.59   & 79.55 & 70.71    & 70.20  & 58.84        & 76.07   \\
Gardening               & 87.76       & 86.46   & 85.42 & 87.50   & 84.11 & 69.53    & 60.42  & 56.77        & 76.69   \\
Household               & 93.12       & 89.68   & 86.77 & 83.33   & 87.57 & 77.78    & 62.96  & 60.05        & 79.89   \\
\midrule[0.05em]
\midrule[0.05em]
Average                 & 77.85       & 77.45   & 75.14 & 74.86   & 76.49 & 63.56    & 60.76  & 51.90        & - \\       
\bottomrule[0.15em]
\end{tabular}
} 
\vspace{-3mm}
\end{table*}

\noindent\textbf{Domain Analysis.}
We also demonstrate the accuracy of models within each domain in Tab.~\ref{tab:comp-uni-domain}.
Specifically, Spatial (mean=57.04) and Plant (mean=59.88) form a high-complexity cluster where depressed scores indicate fundamental architectural bottlenecks in current multimodal frameworks concerning three-dimensional spatial reasoning and fine-grained botanical feature extraction—capabilities demanding robust geometric and symbolic abstraction.
Conversely, quotidian domains such as Household (mean=79.89), Gardening (mean=76.69), and Personal Care (mean=75.93) exhibit performance saturation, reflecting considerable maturity on semantically regular tasks with abundant training corpora.
This performance gradient, from abstract inference to concrete cognition, epitomizes the intrinsic trade-off between domain-generalization capacity and specialized cognitive depth in VLMs, underscoring that heightened generalization may paradoxically constrain proficiency in hard niches.

Contrary to the monotonic capacity-performance hypothesis, several domains witness smaller or ostensibly less-capable models surpassing their larger counterparts.
GPT-4o outperforms Bagel-think in Fluid (71.01 vs. 68.05) despite a 4.14-point aggregate deficit, while UniPic2 ties Ovis-U1 for first place in Animal (79.79), substantially exceeding Bagel-think (77.43). 
These inversion instances challenge the assumption that parameter scale universally dictates domain mastery, likely attributable to domain-specific training data curricula, \eg, GPT-4o’s potential oversampling of fluid dynamics literature, or architectural coincidences, \eg, UniPic2’s patch-embedding strategy serendipitously aligning with animal texture discrimination. 
The phenomena substantiate that, under granular evaluation frameworks, smaller models can achieve asymmetric competitive advantages through strategic optimization, providing empirical justification for deploying specialist architectures alongside generalist systems. 

As shown in Fig.~\ref{fig:display}, images generated by different UMMs under the same prompt vary significantly. 
For instance, in the case placed above, GPT successfully depicts the state of an ungerminated seed and accurately revises the text from `GROW' to `WAIT'. 
In contrast, the images generated by OmniGen2 remain nearly identical across the three rounds. 
Specifically, it fails to revise the text, and the third image contains a large dark region that is presumably attributed to image noise, leading to a lower score.

\subsection{Single Ability}

\noindent\textbf{Understanding Ability.}
Kimi-VL-A3B exhibits unambiguous superiority across all evaluation tiers, securing the highest scores in the generation stage (86.83), interactions stage, and overall performance (75.14). 
Notably, it maintains a distinct performance margin of +1.6 percentage points in generation stage, which underscores its exceptional few-shot visual comprehension capability.

A pronounced bimodal distribution delineates first-tier models from the remainder, with the top three performers (Kimi-VL-A3B, Januspro, and Qwen3-VL-8B) clustering above 66 points.
In contrast, Janus (40.90) and BLIP-VQA (36.07) fail to exceed 60\% of the leading models' scores, underscoring a substantial capability gap between SOTA contemporary architectures and their legacy counterparts.

Degradation rates across stages vary substantially among models.
Kimi-VL-A3B exhibits a 24.15-point decline from the generation stage to the counterfactual stage, while BLIP-VQA shows only a 7.29-point drop.
This observation indicates that although the former maintains absolute performance dominance, the latter demonstrates relatively superior conversational robustness in terms of performance retention.
When assessing understanding ability, the maximum score is bounded by the generation and editing model, \ie GPT-4o and Flux.1-Fill.
However, the comparison between different models still guarantees isotonicity.

\noindent\textbf{Generation Ability.}
The competitive landscape among leading models exhibits extreme convergence, with the top two models, \ie, Unipic (70.56) and OneCat (70.24), separated by a mere 0.32-point differential, and the entire top three spanning only 1.4 points.
This result indicates technological homogenization at the performance ceiling.
\begin{figure}[t]
    \centering
    \includegraphics[width=\linewidth]{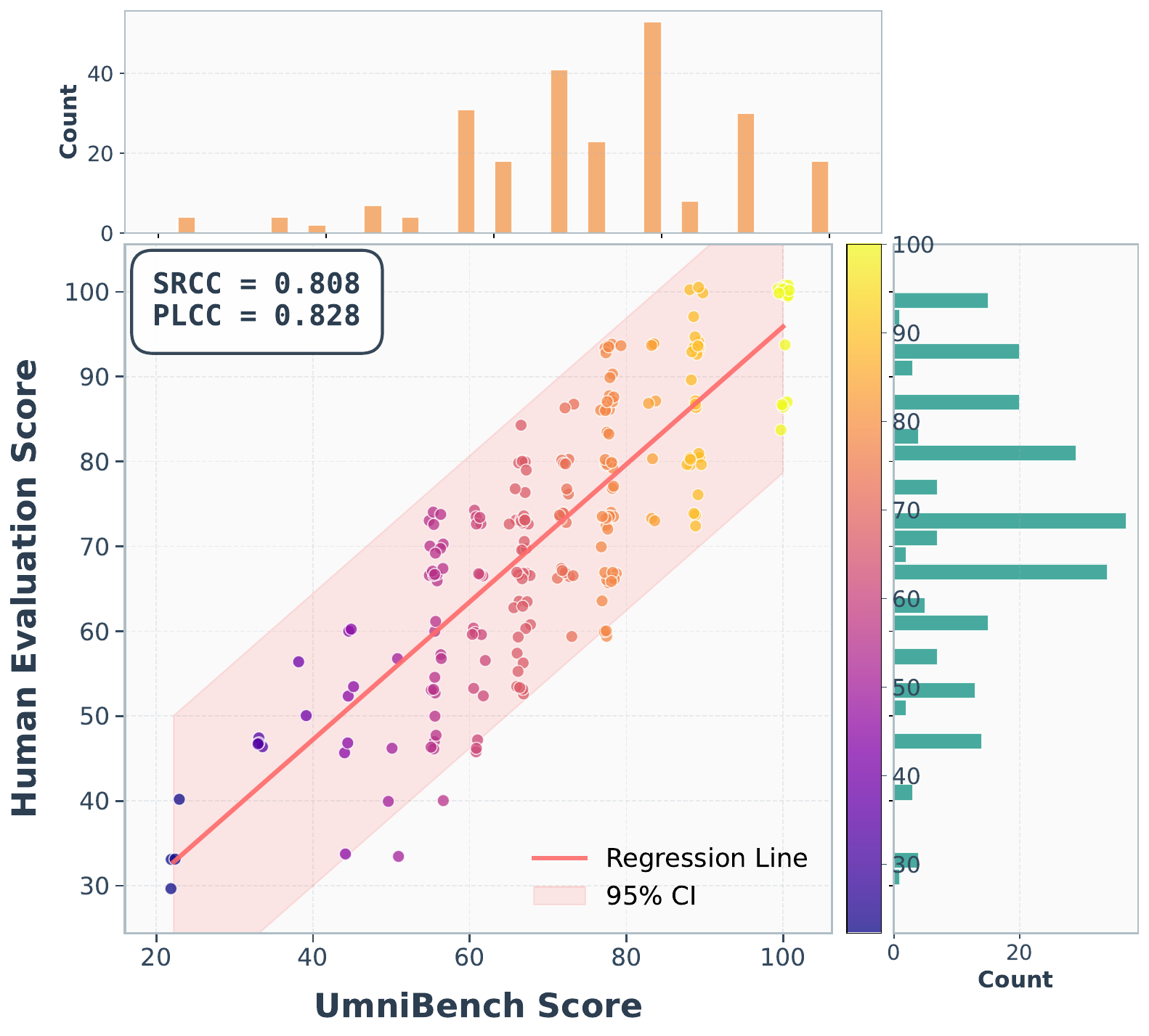}
    \vspace{-8mm}
    \caption{{Correlation with Human Evaluation.} Each spot signifies a case in UmniBench, with the x-axis indicating the UmniBench Score and the y-axis for the human evaluation score.}
    \label{fig:userstudy}
    \vspace{-7mm}
\end{figure}

Emu3 demonstrates exceptional robustness in extended multi-turn generation tasks. 
Although its performance at the generation stage (81.37) is slightly lower than that of JanusPro (84.21), the model accomplishes a performance reversal at counterfactual with a score of 59.38 versus JanusPro’s 55.14. 
Ultimately, it narrows the overall performance disparity to a mere 0.19 points, which attests to its superior multi-turn stability compared to its initial performance.

\noindent\textbf{Editing Ability.}
When assessing editing ability, the score of the generation stage is meaningless, as the model to be assessed is not involved at all.
So we omit the generation score when assessing editing.
Nano-Banana attains comprehensive dominance in editing tasks, securing the top rank in the interaction stage (77.02), and overall performance (77.63). 
This is also in line with the widely recognized understanding among people from industry and academia after extensive testing.
Notably, it maintains a decisive advantage of over 3 points in the interaction stage, which underscores robust performance in complex edit operations.
Qwen-Image-Edit also demonstrates excellent performance in the interaction stage (80.82) but suffers severe degradation in the counterfactual stage (64.09).

\begin{table*}[t]
\centering
\setlength{\tabcolsep}{6pt}
\caption{Single-Ability Comparison. To isolate and evaluate individual capabilities, we adopted a controlled substitution approach: SOTA models were employed to replace the other two target abilities, enabling focused testing of six models across each single ability dimension. }
\label{tab:singel-comp}
\vspace{-3mm}
\resizebox{\linewidth}{!}{
\begin{tabular}{l|l|cccccc}
\toprule[0.15em]
  & Model       & Kimi-VL-A3B~\cite{team2025kimi} & Janus Pro~\cite{chen2025janus}         & Qwen3-VL-8B~\cite{yang2025qwen3} & Emu3~\cite{wang2024emu3}             & Janus~\cite{wu2025janus}        & BLIP-VQA~\cite{li2022blip}     \\
\midrule[0.05em]
\multirow{4}{*}{Und.} & Generation        & 86.83                        & 84.21                    & 85.62                        & 79.89                    & 48.09                & 40.51                \\
                      & Interaction       & 76.47                        & 66.64                    & 64.01                        & 56.69                    & 40.87                & 34.74                \\
                      & Counterfactual    & 62.68                        & 55.14                    & 52.05                        & 47.75                    & 34.10                & 33.22                \\
                      & Overall & 75.14                        & 68.39                    & 66.89                        & 61.10                    & 40.90                & 36.07                \\
\midrule[0.1em]
  & Model       & UniPic2~\cite{wei2025skywork}               & OneCat~\cite{li2025onecat}           & Emu3~\cite{wang2024emu3}                 & Janus Pro~\cite{chen2025janus}         & PixArt-$\alpha$~\cite{chen2023pixart} & Janus~\cite{wu2025janus}     \\
\midrule[0.05em]
\multirow{4}{*}{Gen.} & Generation       & 85.62                        & 84.84                    & 81.37                        & 84.21                    & 77.97                & 81.16                \\
                      & Interaction      & 68.29                        & 68.23                    & 67.50                        & 66.64                    & 68.23                & 66.64                \\
                      & Counterfactual   & 58.56                        & 58.44                    & 59.38                        & 55.14                    & 58.44                & 55.14                \\
                      & Overall & 70.56                        & 70.24                    & 69.20                        & 68.39                    & 68.05                & 67.41                \\
\midrule[0.1em]
 & Model       & Nano-Banana~\cite{comanici2025gemini}  & Qwen-Image-Edit~\cite{wu2025qwen} & FLUX.1-Fill~\cite{flux2024}         & InstructPix2Pix~\cite{brooks2023instructpix2pix} & MagicBrush~\cite{zhang2023magicbrush}  & Step1X-Edit~\cite{liu2025step1x} \\
\midrule[0.05em]
\multirow{3}{*}{Edit}  & Interaction   & 77.02                        & 80.82                    & 68.56                        & 65.39                    & 64.01                & 26.30                \\
                      & Counterfactual   & 70.18                        & 64.09                    & 58.83                        & 51.11                    & 52.05                & 30.80                \\
                      & Overall & 73.68                        & 72.54                   & 63.74                        & 58.32                   & 58.09              & 28.53              \\
\bottomrule[0.15em]
\end{tabular}
} 
\vspace{-3mm}
\end{table*}

\subsection{User Study}
We surveyed to evaluate the correlation between the UmniBench Score and human objective perception. A total of 24 carefully selected volunteers with diverse educational and cultural backgrounds were recruited to rate the images on a 1–5 Likert scale. Specifically, higher scores indicated stronger compatibility between the generated images and their corresponding prompts. For each case, the three ratings were averaged to obtain the human evaluation score. Subsequently, the Pearson Linear Correlation Coefficient (PLCC)~\cite{royal1869proceedings} and Spearman’s Rank Correlation Coefficient (SRCC)~\cite{spearman1961proof} were calculated between these human evaluation scores and the UmniBench scores of each case, where the UmniBench score denotes the accuracy of the model’s responses across all questions included in the case.

As illustrated in Fig.~\ref{fig:userstudy}, a red regression line and its 95\% confidence interval (CI, shaded in light
pink) are presented, with SRCC of 0.808 and PLCC of 0.828. Each data point is color-coded, and the color scale reflects the UmniBench Score of the case, where warmer spots denote a higher score. 
The top histogram depicts the distribution of UmniBench scores, while the right histogram shows the distribution of human evaluation scores, both visualized with bar charts to illustrate the frequency of different score values.
Overall, this visualization reveals a strong consensus between human evaluations and the UmniBench metric. This congruence not only validates the metric’s efficacy in capturing the qualitative nuances of image-prompt alignment but also highlights its potential as a reliable proxy for human assessment in large-scale automated evaluation scenarios.
\vspace{-2mm}
\section{Why is the Score of Counterfactual Low?}
\vspace{-2mm}
The reason behind the low counterfactual score is twofold.

\noindent\textbf{Accumulated Error.}
As the counterfactual stage is the last stage, UMM takes the image generated from the previous stage as input.
If the UMM has already failed on the previous stage, it is not likely that the UMM can provide an accurate result in the counterfactual stage.
Therefore, the success rate is inherently lower than the previous two stages.
Moreover, UMMs usually take well-filtered images during the training process.
However, noise exists in the generated images and will be amplified during the next editing process, leading to louder noise.
These louder noises present as distorted texture, color lump, and irregular colors, significantly influencing the following QAs.

\noindent\textbf{Reasoning Generation.}
All models can hardly perform better than the training set.
When considering generation tasks, the input texts are usually clear and precise attribute descriptions.
This is also a widely used principle to filter the training set.
However, the main content in both the interaction and counterfactual stages involves a reasoning process.
Although the reasoning is not as hard as doing math, this pattern does not appear in the training set, leading to generation or editing failure.
Moreover, the counterfactual stage requires UMMs to replace one of the entities and present new interaction results.
From this perspective, UMMs with token flowing bidirectionally enjoy better reasoning ability.

\vspace{-2mm}
\section{Conclusion}
\vspace{-2mm}
In conclusion, we present UmniBench, a dedicated benchmark for the evaluation of UMMs. 
Leveraging advanced LLM for dataset construction under meticulously curated prompt, we developed a comprehensive collection encompassing 13 primary tags, 195 subtags, and 533  cases. 
This benchmark supports dual-scope evaluation.
It assesses both the overall performance of UMMs and the independent performance of UUMs, as well as single-ability models.
UmniBench exhibits strong holistic coverage and high alignment with human evaluation standards.
Notably, it achieves exceptional efficiency by eliminating dependencies on external tools, LLMs, or human reviewers during evaluation, enabling self-contained capability assessment.
This benchmark is expected to provide valuable support for advancing the development of unified multimodal modeling.

{
    \small
    \bibliographystyle{ieeenat_fullname}
    \bibliography{main}

@String(CVPR= {IEEE Conf. Comput. Vis. Pattern Recog.})

@String(ECCV= {Eur. Conf. Comput. Vis.})

@String(NeurIPS= {Adv. Neural Inform. Process. Syst.})

@String(ICLR = {Int. Conf. Learn. Represent.})

@String(CVPR  = {CVPR})

@String(ECCV  = {ECCV})

@String(ICLR  = {ICLR})

@article{jin2024reasonpix2pix,
  title={Reasonpix2pix: instruction reasoning dataset for advanced image editing},
  author={Jin, Ying and Ling, Pengyang and Dong, Xiaoyi and Zhang, Pan and Wang, Jiaqi and Lin, Dahua},
  journal={arXiv preprint arXiv:2405.11190},
  year={2024}
}

@inproceedings{huang2024smartedit,
  title={Smartedit: Exploring complex instruction-based image editing with multimodal large language models},
  author={Huang, Yuzhou and Xie, Liangbin and Wang, Xintao and Yuan, Ziyang and Cun, Xiaodong and Ge, Yixiao and Zhou, Jiantao and Dong, Chao and Huang, Rui and Zhang, Ruimao and others},
  booktitle={CVPR},
  year={2024}
}

@article{ghosh2023geneval,
  title={Geneval: An object-focused framework for evaluating text-to-image alignment},
  author={Ghosh, Dhruba and Hajishirzi, Hannaneh and Schmidt, Ludwig},
  journal={NeurIPS},
  year={2023}
}

@article{hu2024ella,
  title={Ella: Equip diffusion models with llm for enhanced semantic alignment},
  author={Hu, Xiwei and Wang, Rui and Fang, Yixiao and Fu, Bin and Cheng, Pei and Yu, Gang},
  journal={arXiv preprint arXiv:2403.05135},
  year={2024}
}

@article{huang2025t2i,
  title={T2i-compbench++: An enhanced and comprehensive benchmark for compositional text-to-image generation},
  author={Huang, Kaiyi and Duan, Chengqi and Sun, Kaiyue and Xie, Enze and Li, Zhenguo and Liu, Xihui},
  journal={TPAMI},
  year={2025},
  publisher={IEEE}
}

@article{wu2024conceptmix,
  title={Conceptmix: A compositional image generation benchmark with controllable difficulty},
  author={Wu, Xindi and Yu, Dingli and Huang, Yangsibo and Russakovsky, Olga and Arora, Sanjeev},
  journal={NeurIPS},
  year={2024}
}

@article{wu2025kris,
  title={KRIS-Bench: Benchmarking Next-Level Intelligent Image Editing Models},
  author={Wu, Yongliang and Li, Zonghui and Hu, Xinting and Ye, Xinyu and Zeng, Xianfang and Yu, Gang and Zhu, Wenbo and Schiele, Bernt and Yang, Ming-Hsuan and Yang, Xu},
  journal={arXiv preprint arXiv:2505.16707},
  year={2025}
}

@inproceedings{zeng2025editworld,
  title={Editworld: Simulating world dynamics for instruction-following image editing},
  author={Zeng, Bohan and Yang, Ling and Liu, Jiaming and Xu, Minghao and Zhang, Yuanxing and Wan, Pengfei and Zhang, Wentao and Yan, Shuicheng},
  booktitle={ACM MM},
  year={2025}
}

@article{zhao2025envisioning,
  title={Envisioning beyond the pixels: Benchmarking reasoning-informed visual editing},
  author={Zhao, Xiangyu and Zhang, Peiyuan and Tang, Kexian and Zhu, Xiaorong and Li, Hao and Chai, Wenhao and Zhang, Zicheng and Xia, Renqiu and Zhai, Guangtao and Yan, Junchi and others},
  journal={NeurIPS},
  year={2025}
}

@article{deng2025emerging,
  title={Emerging properties in unified multimodal pretraining},
  author={Deng, Chaorui and Zhu, Deyao and Li, Kunchang and Gou, Chenhui and Li, Feng and Wang, Zeyu and Zhong, Shu and Yu, Weihao and Nie, Xiaonan and Song, Ziang and others},
  journal={arXiv preprint arXiv:2505.14683},
  year={2025}
}

@article{fu2025mme,
    author = {Chaoyou, Fu and Peixian, Chen and Yunhang, Shen and Yulei, Qin and Mengdan, Zhang and Xu, Lin and Jinrui, Yang and Xiawu, Zheng and Ke, Li and Xing, Sun and Yunsheng, Wu and Rongrong, Ji and Caifeng, Shan and Ran, He},
    title ={MME: A Comprehensive Evaluation Benchmark for Multimodal Large Language Models} ,
    journal ={arXiv preprint arXiv:2306.13394} ,
    year = {2023}
}

@inproceedings{liu2024mmbench,
  title={Mmbench: Is your multi-modal model an all-around player?},
  author={Liu, Yuan and Duan, Haodong and Zhang, Yuanhan and Li, Bo and Zhang, Songyang and Zhao, Wangbo and Yuan, Yike and Wang, Jiaqi and He, Conghui and Liu, Ziwei and others},
  booktitle={ECCV},
  year={2024}
}

@inproceedings{yue2024mmmu,
  title={Mmmu: A massive multi-discipline multimodal understanding and reasoning benchmark for expert agi},
  author={Yue, Xiang and Ni, Yuansheng and Zhang, Kai and Zheng, Tianyu and Liu, Ruoqi and Zhang, Ge and Stevens, Samuel and Jiang, Dongfu and Ren, Weiming and Sun, Yuxuan and others},
  booktitle={CVPR},
  year={2024}
}

@inproceedings{yu2024mmvet,
    title={MM-Vet: Evaluating Large Multimodal Models for Integrated Capabilities},
    author={Yu, Weihao and Yang, Zhengyuan and Li, Linjie and Wang, Jianfeng and Lin, Kevin and Liu, Zicheng and Wang, Xinchao and Wang, Lijuan},
    booktitle={ICML},
    year={2024},
}

@inproceedings{lu2024mathvista,
    title={MathVista: Evaluating Mathematical Reasoning of Foundation Models in Visual Contexts},
    author={Lu, Pan and Bansal, Hritik and Xia, Tony and Liu, Jiacheng and Li, Chunyuan and Hajishirzi, Hannaneh and Cheng, Hao and Chang, Kai-Wei and Galley, Michel and Gao, Jianfeng},
    booktitle={ICLR},
    year={2024},
}

@inproceedings{zhang2024mmvp,
    title={MMVP: A Multimodal MoCap Dataset with Vision and Pressure Sensors},
    author={Zhang, He and Ren, Shenghao and Yuan, Haolei and Zhao, Jianhui and Li, Fan and Sun, Shuangpeng and Liang, Zhenghao and Yu, Tao and Shen, Qiu and Cao, Xun},
    booktitle={CVPR},
    year={2024}
}

@article{niu2025wise,
  title={Wise: A world knowledge-informed semantic evaluation for text-to-image generation},
  author={Niu, Yuwei and Ning, Munan and Zheng, Mengren and Jin, Weiyang and Lin, Bin and Jin, Peng and Liao, Jiaqi and Feng, Chaoran and Ning, Kunpeng and Zhu, Bin and others},
  journal={arXiv preprint arXiv:2503.07265},
  year={2025}
}

@article{liu2025step1x,
  title={Step1x-edit: A practical framework for general image editing},
  author={Liu, Shiyu and Han, Yucheng and Xing, Peng and Yin, Fukun and Wang, Rui and Cheng, Wei and Liao, Jiaqi and Wang, Yingming and Fu, Honghao and Han, Chunrui and others},
  journal={arXiv preprint arXiv:2504.17761},
  year={2025}
}

@misc{openai-4o-image-2025,
  author = {OpenAI},
  title = {Introducing 4o image generation},
  year = {2025},
  howpublished = {\url{https://openai.com/index/introducing-4o-image-generation/}},
}

@article{wang2025ovis,
  title={Ovis-U1 Technical Report},
  author={Wang, Guo-Hua and Zhao, Shanshan and Zhang, Xinjie and Cao, Liangfu and Zhan, Pengxin and Duan, Lunhao and Lu, Shiyin and Fu, Minghao and Chen, Xiaohao and Zhao, Jianshan and others},
  journal={arXiv preprint arXiv:2506.23044},
  year={2025}
}

@article{wei2025skywork,
  title={Skywork unipic 2.0: Building kontext model with online rl for unified multimodal model},
  author={Wei, Hongyang and Xu, Baixin and Liu, Hongbo and Wu, Cyrus and Liu, Jie and Peng, Yi and Wang, Peiyu and Liu, Zexiang and He, Jingwen and Xietian, Yidan and others},
  journal={arXiv preprint arXiv:2509.04548},
  year={2025}
}

@article{wu2025omnigen2,
  title={OmniGen2: Exploration to Advanced Multimodal Generation},
  author={Wu, Chenyuan and Zheng, Pengfei and Yan, Ruiran and Xiao, Shitao and Luo, Xin and Wang, Yueze and Li, Wanli and Jiang, Xiyan and Liu, Yexin and Zhou, Junjie and others},
  journal={arXiv preprint arXiv:2506.18871},
  year={2025}
}

@article{li2025onecat,
  title={Onecat: Decoder-only auto-regressive model for unified understanding and generation},
  author={Li, Han and Peng, Xinyu and Wang, Yaoming and Peng, Zelin and Chen, Xin and Weng, Rongxiang and Wang, Jingang and Cai, Xunliang and Dai, Wenrui and Xiong, Hongkai},
  journal={arXiv preprint arXiv:2509.03498},
  year={2025}
}

@article{xin2025lumina,
  title={Lumina-dimoo: An omni diffusion large language model for multi-modal generation and understanding},
  author={Xin, Yi and Qin, Qi and Luo, Siqi and Zhu, Kaiwen and Yan, Juncheng and Tai, Yan and Lei, Jiayi and Cao, Yuewen and Wang, Keqi and Wang, Yibin and others},
  journal={arXiv preprint arXiv:2510.06308},
  year={2025}
}

@article{team2025kimi,
  title={Kimi-vl technical report},
  author={Team, Kimi and Du, Angang and Yin, Bohong and Xing, Bowei and Qu, Bowen and Wang, Bowen and Chen, Cheng and Zhang, Chenlin and Du, Chenzhuang and Wei, Chu and others},
  journal={arXiv preprint arXiv:2504.07491},
  year={2025}
}

@article{chen2025janus,
  title={Janus-pro: Unified multimodal understanding and generation with data and model scaling},
  author={Chen, Xiaokang and Wu, Zhiyu and Liu, Xingchao and Pan, Zizheng and Liu, Wen and Xie, Zhenda and Yu, Xingkai and Ruan, Chong},
  journal={arXiv preprint arXiv:2501.17811},
  year={2025}
}

@article{yang2025qwen3,
  title={Qwen3 technical report},
  author={Yang, An and Li, Anfeng and Yang, Baosong and Zhang, Beichen and Hui, Binyuan and Zheng, Bo and Yu, Bowen and Gao, Chang and Huang, Chengen and Lv, Chenxu and others},
  journal={arXiv preprint arXiv:2505.09388},
  year={2025}
}

@article{wang2024emu3,
  title={Emu3: Next-token prediction is all you need},
  author={Wang, Xinlong and Zhang, Xiaosong and Luo, Zhengxiong and Sun, Quan and Cui, Yufeng and Wang, Jinsheng and Zhang, Fan and Wang, Yueze and Li, Zhen and Yu, Qiying and others},
  journal={arXiv preprint arXiv:2409.18869},
  year={2024}
}

@inproceedings{wu2025janus,
  title={Janus: Decoupling visual encoding for unified multimodal understanding and generation},
  author={Wu, Chengyue and Chen, Xiaokang and Wu, Zhiyu and Ma, Yiyang and Liu, Xingchao and Pan, Zizheng and Liu, Wen and Xie, Zhenda and Yu, Xingkai and Ruan, Chong and others},
  booktitle={CVPR},
  year={2025}
}

@inproceedings{li2022blip,
  title={Blip: Bootstrapping language-image pre-training for unified vision-language understanding and generation},
  author={Li, Junnan and Li, Dongxu and Xiong, Caiming and Hoi, Steven},
  booktitle={ICML},
  year={2022}
}

@article{chen2023pixart,
  title={Pixart-$\alpha $: Fast training of diffusion transformer for photorealistic text-to-image synthesis},
  author={Chen, Junsong and Yu, Jincheng and Ge, Chongjian and Yao, Lewei and Xie, Enze and Wu, Yue and Wang, Zhongdao and Kwok, James and Luo, Ping and Lu, Huchuan and others},
  journal={arXiv preprint arXiv:2310.00426},
  year={2023}
}

@article{comanici2025gemini,
  title={Gemini 2.5: Pushing the frontier with advanced reasoning, multimodality, long context, and next generation agentic capabilities},
  author={Comanici, Gheorghe and Bieber, Eric and Schaekermann, Mike and Pasupat, Ice and Sachdeva, Noveen and Dhillon, Inderjit and Blistein, Marcel and Ram, Ori and Zhang, Dan and Rosen, Evan and others},
  journal={arXiv preprint arXiv:2507.06261},
  year={2025}
}

@article{wu2025qwen,
  title={Qwen-image technical report},
  author={Wu, Chenfei and Li, Jiahao and Zhou, Jingren and Lin, Junyang and Gao, Kaiyuan and Yan, Kun and Yin, Sheng-ming and Bai, Shuai and Xu, Xiao and Chen, Yilei and others},
  journal={arXiv preprint arXiv:2508.02324},
  year={2025}
}

@inproceedings{brooks2023instructpix2pix,
  title={Instructpix2pix: Learning to follow image editing instructions},
  author={Brooks, Tim and Holynski, Aleksander and Efros, Alexei A},
  booktitle={CVPR},
  year={2023}
}

@inproceedings{zhang2023magicbrush,
  title={Magicbrush: A manually annotated dataset for instruction-guided image editing},
  author={Zhang, Kai and Mo, Lingbo and Chen, Wenhu and Sun, Huan and Su, Yu},
  booktitle={NeurIPS},
  year={2023}
}

@misc{flux2024,
    author={Black Forest Labs},
    title={FLUX},
    year={2024},
    howpublished={\url{https://github.com/black-forest-labs/flux}},
}

@inproceedings{das2024exams,
  title={Exams-v: A multi-discipline multilingual multimodal exam benchmark for evaluating vision language models},
  author={Das, Rocktim and Hristov, Simeon and Li, Haonan and Dimitrov, Dimitar and Koychev, Ivan and Nakov, Preslav},
  booktitle={ACL},
  year={2024}
}

@article{pan2025ice,
  title={Ice-bench: A unified and comprehensive benchmark for image creating and editing},
  author={Pan, Yulin and He, Xiangteng and Mao, Chaojie and Han, Zhen and Jiang, Zeyinzi and Zhang, Jingfeng and Liu, Yu},
  journal={arXiv preprint arXiv:2503.14482},
  year={2025}
}

@inproceedings{li2025llava,
  title={Llava-st: A multimodal large language model for fine-grained spatial-temporal understanding},
  author={Li, Hongyu and Chen, Jinyu and Wei, Ziyu and Huang, Shaofei and Hui, Tianrui and Gao, Jialin and Wei, Xiaoming and Liu, Si},
  booktitle={CVPR},
  year={2025}
}

@book{royal1869proceedings,
  title={Proceedings of the Royal Society},
  author={Royal Society (London)},
  volume={17},
  year={1869}
}

@article{spearman1961proof,
  title={The proof and measurement of association between two things.},
  author={Spearman, Charles},
  year={1961},
  publisher={Appleton-Century-Crofts}
}
}
\end{document}